\documentclass[letterpaper, 10 pt, conference]{ieeeconf}  

\IEEEoverridecommandlockouts                              

\usepackage{amsmath}
\usepackage{siunitx}
\usepackage{graphicx}
\usepackage{gensymb}
\usepackage{amsmath}
\usepackage{mathtools}

\DeclareMathOperator{\atantwo}{atan2}
\DeclarePairedDelimiter\abs{\lvert}{\rvert}

\title{\LARGE \bf
DRBA: Dynamic Robotic Balance Assistant - An assist-as-needed gait and balance rehabilitation robot for versatile training
}

\author{Yifan Wang$^{*1}$, Lei Li$^{*2}$, Youlong Wang$^{1}$, Chengyuan Yang$^{1}$, Sherwin Stephen Chan$^{1}$, Jiaye Chen$^{1}$, \\Xiaoyue Yan$^{1}$,Hao Wang$^{2}$, Xuesheng Gong$^{3}$, Jun Lin$^{3}$, Hongping Hu$^{3}$, and Wei Tech Ang$^{1}$
\thanks{*indicates equal contribution}
\thanks{$^{1}$Yifan Wang, Youlong Wang, Chengyuan Yang, Sherwin Stephen Chan, Jiaye Chen, Xiaoyue Yan and Wei Tech Ang are with School of Mechanical and Aerospace Engineering,
        Nanyang Technological University, 639798 Singapore
        {\tt\small Corresonding email: ywang114@e.ntu.edu.sg}}%
\thanks{$^{2}$Lei Li, and Hao Wang are with Guangdong Zhongxin Intelligent Rehabilitation Research Institute, Foshan, China 528200}%
\thanks{$^{3}$Xuesheng Gong, Jun Lin and Hongping Hu are with Guangdong Jianxiang Hospital Group, Foshan, China 528200}%
}

\begin{document}

\maketitle
\thispagestyle{empty}
\pagestyle{empty}

\begin{abstract}

The decline of human balance control due to aging and pathological conditions increases fall risk, a major concern in geriatric care and rehabilitation. Gait training is essential for balance recovery, enhancing walking ability and postural control. However, existing overground robotic gait trainers have limitations: body weight support systems are bulky and impractical for daily use, while end-effector-based systems often compromise transparency, altering natural gait dynamics. This paper presents the Dynamic Robotic Balance Assistant (DRBA), a novel gait trainer providing assist-as-needed body weight and balance support for various training scenarios. DRBA integrates a 3-degree-of-freedom (3-DoF) robotic arm for pelvic support with flexible motion, a compact sit-to-stand assistance module, and user-following and fall detection algorithms to ensure minimal interference and responsive support. Experimental results demonstrated high transparency, with minimal impact on natural gait dynamics. A patient trial with nine elderly patients with varying medical conditions and balance impairments (ranging from severe to mild) further validated DRBA’s effectiveness. The results showed that DRBA-assisted training increased step length and walking speed compared to therapist-assisted gait training. Additionally, DRBA enabled users to perform tasks beyond their unaided ability, expanding rehabilitation possibilities. These findings highlight DRBA’s potential to enhance rehabilitation outcomes by facilitating higher training intensity and enabling task-oriented exercises.
\end{abstract}

\section{INTRODUCTION}



Human balance control naturally declines with age and is further affected by neurological disorders (e.g., stroke, spinal cord injury, Parkinson’s disease), musculoskeletal conditions (e.g., chronic low back pain, scoliosis, amputation), and vestibular deficits (e.g., benign paroxysmal positional vertigo). These impairments often lead to reduced proprioception and coordination, increasing the risk of falls, which poses significant challenges for elderly and impaired individuals undergoing rehabilitation \cite{nyberg1995patient}. Given that balance control is critical for activities of daily living (ADL) \cite{keenan1984factors} and is a strong predictor of independent living \cite{desrosiers2002predictors}, its deterioration significantly impacts both physical function and quality of life. Additionally, impaired balance is a leading cause of self-perceived disability post-rehabilitation \cite{lin2001predicting}, underscoring the need for effective interventions to restore balance and mobility.

Gait training is a widely recognized approach for balance recovery \cite{pollock2000balance, mehrholz2017treadmill, jezernik2003robotic}, as it improves both walking ability and postural control by integrating balance mechanisms into locomotion. Over the past decade, overground gait and balance trainers have been developed to enhance rehabilitation outcomes, reduce caregiver burden, and improve safety during gait training
 
Overground gait trainers generally fall into two categories: body weight support systems and end-effector-based systems. Body weight support systems, such as Andago \cite{marks2019andago}, may include a mobile base \cite{marks2019andago,leme2021socially} or function without one \cite{vallery2013multidirectional,frey2006novel}. These systems provide continuous partial or full body weight support through a suspended harness mechanism, reducing physical strain on the patient during rehabilitation. However, their bulky structural design and large footprint limit maneuverability, requiring spacious training environments, making them primarily suitable for hospital and rehabilitation institution use.

End-effector-based systems, such as KineAssist \cite{patton2008kineassist}, integrate a robotic arm for user interaction coupled with a mobile base that moves with the user. Compared to body weight support systems, these robots are more compact and flexible, making them suitable for outdoor and community-based rehabilitation. However, existing end-effector-based systems often struggle with transparency issues in their physcial human robot interface (pHRI) design, imposing unintended mechanical constraints on the user’s natural movement, ultimately reducing gait training effectiveness. Early robotic gait trainers, such as KineAssist \cite{patton2008kineassist} and SoloWalk \cite{morbi2014design}, utilized rigid robotic arms for support and protection. However, these rigid structures transmitted inertial forces from the mobile base to the user, distorting gait dynamics and altering natural gait patterns. Later systems attempted to mitigate these inertial effects using smooth tracking controllers based on human-robot interaction forces \cite{mun2014design,aguirre2021omnidirectional}, but gait alterations were still observed \cite{mun2017biomechanical}. Additionally, their rigid mechanical constraints restricted user movement, limiting engagement in diverse rehabilitation tasks and ADL activities. 

Recent efforts have introduced compact robotic gait trainers designed for home and community use \cite{li2023mobile,wang2023graceful}. These systems integrate passive parallel robotic arms to decouple mobile base dynamics from the user, improving gait training transparency. However, their integration with electrically powered wheelchairs, intended for daily transportation, compromises pHRI transparency by restricting free movement. Additionally, the large inertia of the mobile base induces a pulling-pushing effect, further disrupting natural gait patterns during training \cite{li2023mobile}.

Despite advancements, existing robotic gait trainers still face trade-offs among mobility, transparency, and adaptability. Body weight support systems, though effective, are bulky and limited to clinical settings, while end-effector-based systems constrain natural gait. Even newer compact designs restrict free movement and introduce unwanted dynamics. These challenges underscore the need for a lightweight, adaptable, and transparent gait trainer that preserves natural movement across diverse rehabilitation environments.

In this paper, we introduce \textbf{DRBA} – Dynamic Robotic Balance Assistant, a rehabilitation-oriented robotic platform designed to provide assist-as-needed body-weight and balance support for overground gait training. Compared with prior mobile robotic balance assistants and our previous MRBA-related work \cite{li2023mobile,wang2023graceful}, DRBA is specifically designed for rehabilitation rather than daily-living mobility assistance. The system integrates: (i) a low-inertia compliant pelvic interface that allows 3-DoF planar relative motion between the user and the robot, reducing the transmission of mobile-base inertia; (ii) an integrated sit-to-stand mechanism that supports seamless transitions between seated transfer and gait training; and (iii) a user-following and fall-intervention architecture that provides assist-as-needed support during rehabilitation tasks. Together, these features aim to deliver safe, transparent, and adaptable assistance while preserving natural pelvic motion.

By reducing balance demands and increasing users’ confidence during training, DRBA has the potential to facilitate higher-intensity, longer-duration, and more task-oriented rehabilitation. Preliminary transparency evaluation suggested limited interference with lower-limb kinematics during walking, while pilot studies involving nine elderly patients with balance impairments demonstrated significant improvements in step length and walking speed during DRBA-assisted walking compared with therapist-assisted walking. Furthermore, DRBA enabled users to perform rehabilitation activities beyond their unaided physical capacity, supporting functional and challenge-based training. These findings demonstrate the immediate assistive benefits of DRBA and highlight its potential to improve rehabilitation outcomes through increased training dosage and engagement. However, whether these benefits translate into long-term recovery remains to be established through longitudinal clinical studies.

Our key contributions include: 
\begin{itemize} 
    \item \textbf{Dynamic Robotic Balance Assistant (DRBA):} A novel robotic gait trainer that provides assist-as-needed body weight and balance support, preserving natural pelvic motion within a compact and mobile form factor, making it suitable for rehabilitation centers, community care settings, and outdoor use.
    \item \textbf{Patient Trials:} We conducted patient studies with nine patients with various medical conditions and balance impairments (BBS: 8–50). Results demonstrated significant improvements in gait parameters and enhanced safety, enabling higher-intensity exercises and a broader range of rehabilitation activities.
\end{itemize}

\section{SYSTEM OVERVIEW}
\begin{figure*}[thpb]
\centering
\includegraphics[width=\linewidth]{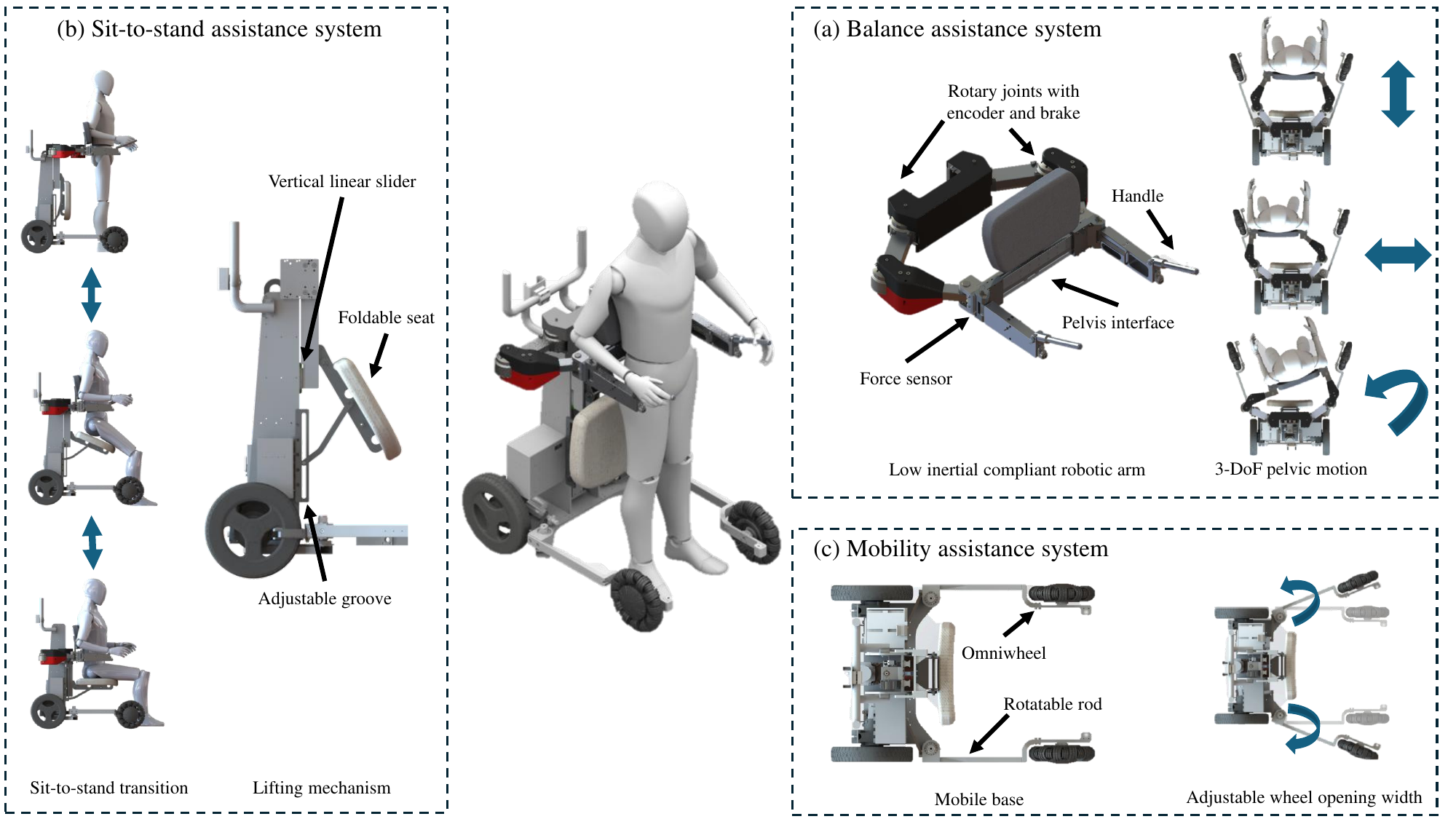}
\caption{System Overview of DRBA. (a) Balance assistance system: monitors the user’s state and provides adaptive balance support through a low inertia compliant robotic arm, enabling 3-DoF pelvic motion. (b) Sit-to-stand assistance system: assists users in transitioning between sitting and standing, featuring a lifting mechanism with a vertical linear slider and a foldable seat. (c) Mobility assistance system: allows the robot to move together with the user via a mobile base with two omniwheels mounted on rotatable rods, enabling adjustable wheel spacing to accommodate different walking needs.}
\label{fig:DRBA_overview}
\end{figure*}

DRBA comprises three subsystems: 1) Balance assistance system, 2) Sit-to-stand assistance system and 3) Mobility assistance system to assist and enhance users' balance and mobility training. An overview of the DRBA system is presented in Fig. \ref{fig:DRBA_overview}. The control architecture integrates a user-following controller and an instability detection and fall intervention mechanism, enabling the robot to synchronously accompany the user and provide timely balance support when instability is detected.

\subsection{Balance assistance system}
The balance assistance system continuously monitors the user’s state and provides support as needed. It features a low-inertia compliant robotic arm that wraps around the user’s pelvis. Similar to systems like KineAssist \cite{patton2008kineassist} and SoloWalk \cite{morbi2014design}, this design targets the pelvis—located near the center of mass (CoM)—to deliver balance interventions more effectively. However, unlike these systems, which use rigid robotic arms, our design incorporates a passive, compliant structure that allows 3-DoF planar relative motion of the user's pelvis. This critical design choice decouples the robot’s main body inertia from the user during overground gait training, minimizing impedance forces when balance assistance is not required. The robotic arm primarily enables pelvic motion in the transverse plane, but our pHRI design incorporates elastic cushions and adjustable belts to provide a unique yet limited degree of freedom in other directions, such as pelvic tilt, which depends on the deformation of these materials. This flexible design enables users to perform functional and ADL tasks, such as kicking a ball or wearing pants, which require vertical CoM manipulation.

The robotic arm consists of six rotary joints with a pelvis interface. Four of these joints are equipped with encoders and brakes. The encoders measure the position and orientation of the interface, enabling continuous monitoring of the relative position between the user and the robot. The brakes provide balance assistance and fall intervention by locking the robotic arm to stabilize the user in times of instability. Additionally, two force sensors are symmetrically mounted on the pelvis interface to detect the user’s body weight, which is transferred through the safety belt. Additionally, detachable handles are attached to the pelvis interface, providing users with limited mobility an optional self-support mechanism as needed.

\subsection{Sit-to-stand assistance system}

The sit-to-stand assistance system enables safe, seamless transfers from hospital beds to rehabilitation gyms, reducing therapist workload. It features a vertical linear slider powered by a linear driving module. A foldable seat is attached to both the slider and the base of the robot. The seat’s linkage joint moves within a sliding groove on the base, which has an adjustable limit to accommodate different seat heights based on user needs. The transition from sitting to standing occurs in two stages: i) folding stage: the seat linkage operates as a four-bar linkage, shifting the seat from a horizontal to a vertical position as it folds onto the lifting mechanism while the slider ascends. ii) rising stage: once the seat reaches the vertical position, it remains upright and continues to move with the slider until the joint in the groove reaches its top limit. Separately, the 3 DoF robotic arm is also mounted on the slider which allows the user to adjust an appropriate height for standing and walking.

\subsection{Mobility assistance system}
The mobility assistance system allows the robot to move together with the user, providing support in time across various training environments. It features a mobile base equipped with two actuated rear wheels and two passive front omniwheels. The use of omniwheels enhances the robot’s ability to transparently track the user’s movements, accommodating both rotational and lateral motion of the pelvis during forward walking. Each omniwheel is independently connected to the base via a separate connecting rod, which can rotate around its joint on the base to adjust the wheel’s opening width. This adjustable design offers greater flexibility for the user, enabling a more comfortable walking space during training while also allowing easy passage through doorways when transitioning between different environments and facilities.
\subsection{Control architecture}
\subsubsection{User following control}
\begin{figure}[thpb]
\centering
\includegraphics[width=\linewidth, height=0.2\textheight]{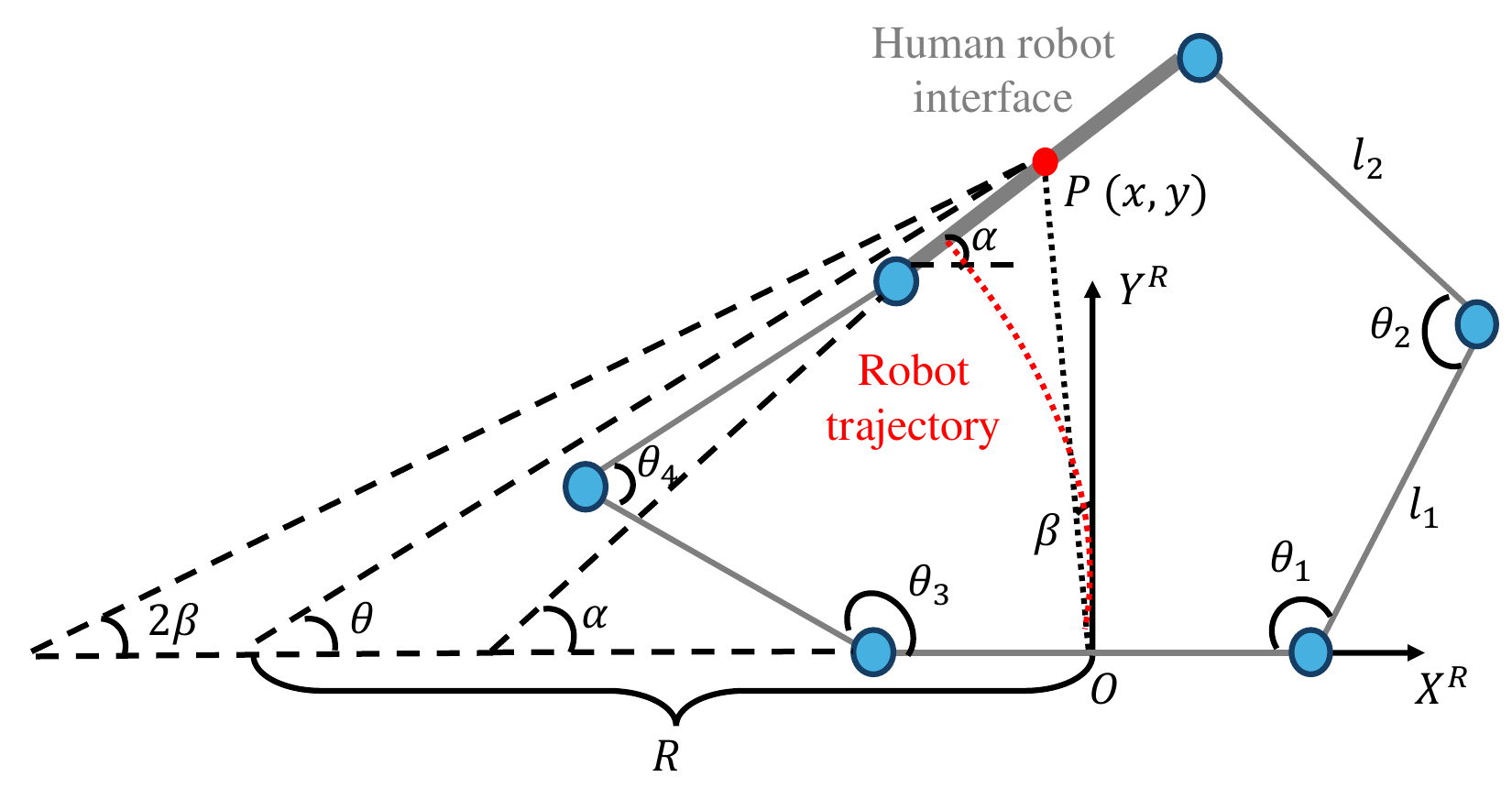}
\caption{Schematic diagram of the user following control algorithm. $X^R\text{-}O\text{-}Y^R$ denotes the robot frame. Joint angles $\theta_1, \theta_2, \theta_3$ and $\theta_4$ are measured by encoders. The lengths of the proximal and distal links of the robotic arm on one side are denoted by $l_1$ and $l_2$. Point $P$ represents the center of the human-robot interface, with coordinates $(x, y)$ defined in the robot’s frame and the orientation of the interface is given by $\alpha$. $\beta$ is the angle between $\vec{OP}$ and $Y^R$ axis. $\theta$ and $R$ represent the actual turning angle and turning radius of the robot.}
\label{fig:User_following}
\end{figure}

DRBA follows the user by tracking the state of the user’s pelvis, which is connected to the robotic arm interface, as shown in Fig \ref{fig:User_following} . The pelvis state $(x,y,\alpha)$, defined by the center position and orientation of the interface, is determined using measurements from the robotic arm’s encoders:

\begin{equation}
\begin{split}
    x = \frac{l_1\cos(\pi-\theta_1)+l_2\cos(\pi-\theta_1-\theta_2)}{2}\\
    + \frac{l_1\cos\theta_3+l_2\cos(\theta_3+\theta_4-\pi)}{2}\quad\quad
\end{split}
\end{equation}
\begin{equation}
\begin{split}
    y = \frac{l_1\sin(\pi-\theta_1)+l_2\sin(\pi-\theta_1-\theta_2)}{2}\\
    + \frac{l_1\sin\theta_3+l_2\sin(\theta_3+\theta_4-\pi)}{2}\quad\quad
\end{split}
\end{equation}
\begin{equation}
\begin{split}
    \alpha = \atantwo (l_1(\sin(\pi-\theta_1)-\sin\theta_3)+\quad\quad\quad\quad\quad\quad\\l_2(\sin(\pi-\theta_1-\theta_2)-\sin(\theta_3+\theta_4-\pi),\\
    l_1(\cos(\pi-\theta_1)-\cos\theta_3)+\quad\quad\quad\quad\quad\quad\\l_2(\cos(\pi-\theta_1-\theta_2)-\cos(\theta_3+\theta_4-\pi))
\end{split}
\end{equation}
where $\theta_1, \theta_2, \theta_3$ and $\theta_4$ are the joint angles measured by the encoders, and $l_1$ and $l_2$ represent the lengths of the proximal and distal links of the robotic arm on one side respectively.

The user-following control ensures the robot maintains a safe distance from the user while aligning its movement with the user’s direction. To prevent collisions, the robot stops if the distance falls below a set safety threshold. If the user moves forward and exceeds this threshold, the robot follows to catch up. Additionally, it continuously adjusts its orientation to stay aligned with the user’s heading.

Due to the non-holonomic nature of the differential drive, the robot moves along a turning radius $R$ and a turning angle $\theta$, as illustrated in Fig \ref{fig:User_following}. When the user turns with DRBA, they may either rotate their pelvis or take side steps. These two turning preferences result in two possible turning angles for the robot respectively: $\alpha$ and $2\beta$, where $\beta=\atantwo(y,x)$. The robot’s actual turning angle  $\theta$  is a weighted combination of these two angles:
\begin{equation}
    \theta=k_1\alpha+2k_2\beta
\end{equation}
Here, $k_1$ and $k_2$ are tunable parameters that can be adjusted based on the user’s turning preference, with the constraint $k_1 + k_2 = 1$.

The turning radius is calculated as:
\begin{equation}
    R = \frac{\abs{\Vec{OP}}{}\sin(\frac{\pi}{2}-\theta+\beta)}{\sin\theta}
\end{equation}
where $\abs{\Vec{OP}}{}=\sqrt{x^2+y^2}$ represents the distance from the robot to the user.

The desired linear velocity $v$ of the robot is proportional to the arc length determined by the turning radius and angle:
\begin{equation}
    v = k_pR\theta
\end{equation}
where $k_p$ is the tunable proportional gain.

Accordingly, the left and right wheel speeds, $\omega_l$ and $\omega_r$, can be computed respectively as:
\begin{equation}
    \omega_l=\frac{k_p\theta(R-\frac{L}{2})}{r},\quad\omega_r=\frac{k_p\theta(R+\frac{L}{2})}{r}
\end{equation}
where $L$ is the distance between the two wheels and $r$ is the wheel radius.

\subsubsection{Instability detection and fall intervention}
The instability detection and fall intervention algorithm aims to identify potential fall events during gait training and determine the appropriate moment to intervene, ensuring the user's safety. DRBA continuously monitors the user’s stability by tracking body weight transferred to the robotic arm via two force sensors.

The core principle for fall detection is to identify significant changes in supporting forces within a short time frame. Using a sliding window of size $T$, the force measurements from the left and right sensors are denoted as $F_l$ and  $F_r$ respectively. The changes in forces during this window are calculated as: $\Delta F_l=F_l(T)-F_l(0)$ and $\Delta F_r=F_r(T)-F_r(0)$. The state of the user $S_{user}$ is determined based on these force changes accordingly:
\begin{gather}
    S_{user} = 
    \begin{cases}
    \text{Lateral fall} & \text{if }\Delta F_{l} > \epsilon_1\text{ or }\Delta F_{r} > \epsilon_1\\
    \text{Downward fall}  & \text{if }\Delta F_{l}+\Delta F_r > \epsilon_2\\
    \text{Stable walking}  & \text{otherwise}
    \end{cases}
\end{gather}
Here, $\epsilon_1$ and $\epsilon_2$ represent adjustable thresholds for detecting lateral falls and downward falls respectively. These parameters can be fine-tuned during gait training to align with the user’s physical abilities and rehabilitation intensity of challenge-based tasks. Once a fall is detected, the robotic arm’s brakes are immediately engaged to lock the robotic arm to provide balance support and prevent the fall. After the user regains stability, they can unlock the robotic arm using a designated button to resume gait training.

\section{EXPERIMENT}
\begin{figure}[thpb]
\centering
\includegraphics[width=\linewidth, height=0.25\textheight]{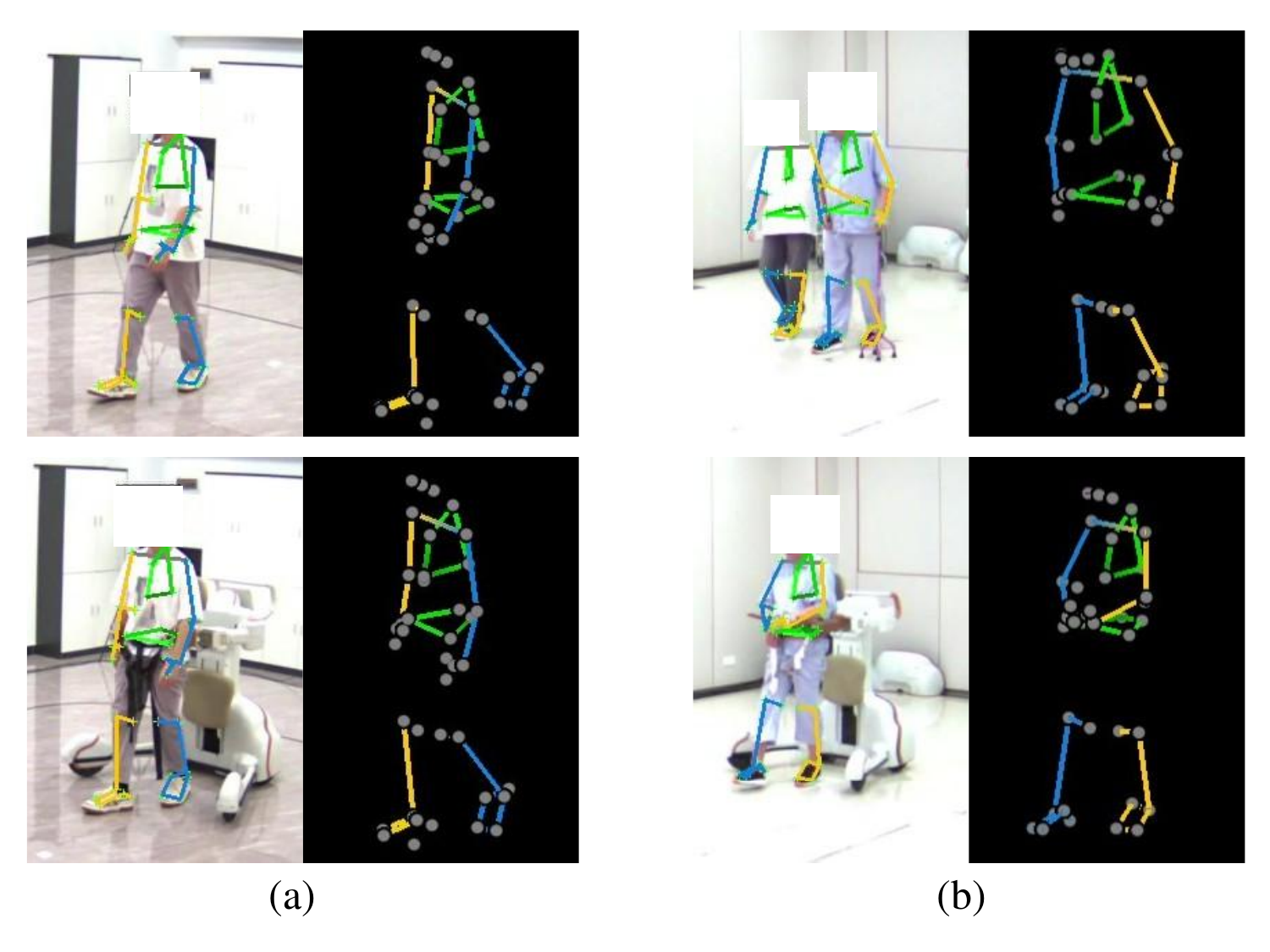}
\caption{Experiment setup. (a) Transparency evaluation. Upper: healthy subject free walking. Lower: healthy subject walking with DRBA. (b) Effectiveness evaluation. Upper: patient assisted by a therapist. Lower: patient assisted by DRBA.}
\label{fig:Experiment_settings}
\end{figure}

The evaluation of DRBA was conducted from two perspectives: (i) transparency and (ii) effectiveness. The transparency evaluation assesses the extent to which DRBA interferes with users by examining whether DRBA alters the natural gait of healthy individuals who do not require assistance. The effectiveness evaluation examines whether DRBA can improve users’ gait and enhance rehabilitation training. This evaluation involves nine elderly patients with balance impairments as seen in Table \ref{tb:SubjectInfo} to determine if DRBA supports effective gait training. All experiments utilize a markerless motion capture system \cite{jatesiktat2024anatomical} with a sampling rate of \SI{30}{\hertz} for gait parameter analysis as shown in Fig \ref{fig:Experiment_settings}. The study protocol was approved under JK-2023A-05 by the Institutional Review Board of Guangdong Jianxiang Hospital Group.

\subsection{Transparency evaluation}
A healthy male participant with no known gait or balance impairments was recruited. The subject completed two experimental conditions designed to compare natural walking patterns with those observed when using DRBA: (1) free walking - the participant was instructed to walk a straight path independently at his preferred walking speed. To capture a comprehensive profile of natural gait variability, the subject completed five walking trials, deliberately varying his speed from slow to his maximum comfortable pace across the trials. (2) walking with DRBA -  the participant completed an additional five walking trials along the same path while using DRBA. Prior to data collection, the participant was provided with sufficient time to be familiarized with DRBA, ensuring an adequate adaptation period to minimize learning effects.

\subsection{Effectiveness evaluation}
Nine elderly patients with various medical conditions and balance impairments were recruited for the trial. Their balance impairments were evaluated using the Berg Balance Scale (BBS) \cite{berg1989measuring}. Participants were categorized into three functional groups according to their BBS scores:
(i) Severe balance impairment ($0$-$20$) - 3 subjects.
(ii) Moderate balance Impairment ($21$-$40$) - 5 subjects. (iii) Mild to no balance impairment ($41$-$50$) - 1 subject. Table \ref{tb:SubjectInfo} summarizes the demographic and clinical characteristics of the recruited subjects.

\begin{table}[hb]
\begin{center}
\caption{Demographic and Clinical Characteristics of Subjects}\label{tb:SubjectInfo}
\begin{tabular}{ccccc}
\hline
Subject & Gender & Age & Disease type & BBS score\\
\hline
1 & Female & 84 & Stroke & 26 \\
2 & Male & 87 & Stroke & 50 \\
3 & Female & 69 & Stroke & 23 \\
4 & Female & 76 & Parkinson \& spondylolisthesis & 8 \\
5 & Male & 75 & Stroke & 40 \\
6 & Female & 76 & Stroke & 22 \\
7 & Female & 79 & Stroke & 35 \\
8 & Female & 76 & Stroke \& Parkinson & 9 \\
9 & Male & 88 & Stroke & 20 \\
\hline
\end{tabular}
\end{center}
\end{table}

Patients underwent two distinct walking trials designed to evaluate DRBA’s impact on their gait performances: 

(1) 10-meter walk assisted by therapist. In this trial, subjects were instructed to walk a straight 10-meter path as quickly as possible while utilizing a crutch and receiving manual assistance from a therapist when necessary. This condition served as a baseline for traditional, therapist-assisted walking without robotic intervention.

(2) 10-meter walk assisted by DRBA. The subjects completed the 10-meter walk under DRBA-assisted condition without direct assistance from the therapist. Prior to data collection, participants received guided training sessions to familiarize themselves with DRBA’s operation. Therapists facilitated the initial walking sessions, progressively reducing their level of assistance as participants adapted to DRBA. The level of robotic assistance was tuned individually to suit each participant’s specific needs.

\section{RESULTS AND DISCUSSION}
\subsection{Transparency}
\begin{figure}[thpb]
\centering
\includegraphics[width=\linewidth, height=0.2\textheight]{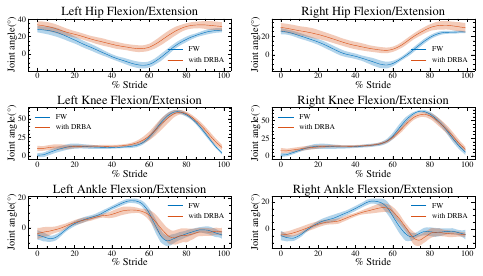}
\caption{Comparison of Lower Limb Joint Kinematics in free Walking and DRBA-Assisted walking. The graphs depict sagittal plane joint angles for the hip, knee, and ankle over a complete gait cycle. 'FW' represents free walking without the robot, while 'with DRBA' indicates walking with robotic assistance. Shaded areas denote the standard deviation of joint angles across multiple gait cycles.}
\label{fig:Transparency}
\end{figure}

The lower limb joint kinematics of the healthy subject in the sagittal plane were analyzed to evaluate the interference introduced by DRBA. Kinematic data from four complete gait cycles were selected from each trial under both free walking and DRBA-assisted walking conditions for comparison. The resulting joint trajectories are presented in Fig \ref{fig:Transparency}.

The figure reveals a reduced range of motion (ROM) in the hip joint during DRBA-assisted walking, particularly characterized by a noticeable decrease in hip extension. Although DRBA was designed to provide sufficient space in the lower limb region to accommodate the user’s natural range of motion, healthy subjects tend to consciously reduce their step length and hip extension when adapting to DRBA. This precautionary adjustment arises from an attempt to avoid potential collisions with the robot, especially since the user cannot directly perceive the device positioned behind them. Additionally, a mismatch in DRBA tuning may also contribute to the reduced hip extension. DRBA is primarily tuned for patients with a typical maximum walking speed below $0.8$ m/s. This is supported by gait speed analysis, which showed the subject slowing down to $0.822 \pm 0.074$ m/s to adapt to the robot.

For the knee joint, no significant differences in the range of motion were observed between free walking and DRBA-assisted walking, indicating minimal interference from DRBA at this joint. However, in the ankle joint, a slight reduction in ankle dorsiflexion was noted just before toe-off ( $60\%$ of the gait cycle). This minor alteration appears to be a secondary effect of the reduced hip extension and shorter step length.

Furthermore, the subject maintains good gait symmetry while walking with DRBA. The left and right step lengths during DRBA-assisted walking are $0.440 \pm 0.034$ m and $0.454 \pm 0.032$ m, respectively, while the left and right step times are $0.541 \pm 0.037$ s and $0.546 \pm 0.035$ s. Additionally, no significant inter-cycle gait variations were observed when comparing DRBA-assisted walking to free walking, as demonstrated by the shaded regions of the joint trajectories in Figure \ref{fig:Transparency}. This suggests that the user maintains consistent gait stability while walking with DRBA, indicating that the robot introduces minimal interference and does not distort natural gait dynamics.


In summary, the preliminary transparency evaluation suggested that DRBA preserved the main characteristics of the subject’s lower-limb kinematics, particularly at the knee joint, while only minor deviations were observed at the hip and ankle. These deviations were likely related to the subject’s conscious adaptation to the robot and the patient-oriented tuning of DRBA, rather than direct mechanical constraint alone. However, the evaluation involved only one healthy participant and should therefore be interpreted as an initial feasibility result rather than definitive validation of transparency. Future work will include larger cohorts of healthy participants and patients to systematically evaluate transparency across different walking speeds, body sizes, and gait patterns.

\subsection{Effectiveness}
\begin{figure}[thpb]
\centering
\includegraphics[width=\linewidth, height=0.2\textheight]{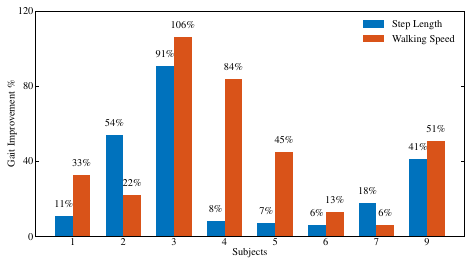}
\caption{Gait Improvement in DRBA-Assisted vs. therapist-assisted training. The graph presents the percentage improvement in step length (blue bars) and walking speed (red bars) for all subjects during DRBA-assisted training compared to traditional therapist-assisted training.}
\label{fig:Gait_improvement}
\end{figure}

The step length and walking speed of the nine elderly patients were analyzed to assess whether DRBA-assisted walking improves gait performance compared to conventional therapist-assisted walking. The improvement in step length and walking speed across all subjects is presented in Figure \ref{fig:Gait_improvement}. During the trial, Subject 8 was unable to walk independently with DRBA despite substantial facilitation from the therapist. This was attributed to the subject’s lack of gait training over the previous two years, which resulted in severely deteriorated balance perception and a high dependence on therapist assistance when attempting to walk with DRBA. Consequently, Subject 8 was excluded from the quantitative gait analysis because the influence of therapist assistance could not be eliminated.

To evaluate whether the observed improvements were statistically significant, paired t-tests were performed between the therapist-assisted and DRBA-assisted conditions for the remaining eight subjects. The results showed a significant increase in step length from $0.187 \pm 0.073$ m to $0.231 \pm 0.078$ m ($t(7)=3.18$, $p=0.015$), corresponding to an average improvement of 23.5\%. Walking speed also increased significantly from $0.214 \pm 0.090$ m/s to $0.290 \pm 0.090$ m/s ($t(7)=4.73$, $p=0.002$), representing an average improvement of 35.7\%. Furthermore, all analyzed subjects demonstrated improvements in both step length and walking speed during DRBA-assisted walking compared with therapist-assisted walking.

Although gait performance improved under DRBA assistance, our patient trials did not provide statistically significant evidence that the BBS score predicts the magnitude of gait improvement achieved with DRBA. Pearson correlation analysis yielded a coefficient of $-0.25$ ($p=0.553$), suggesting a weak negative relationship that was not statistically significant. To further explore which users benefit most from DRBA-assisted gait training, we conducted a case-by-case analysis incorporating Manual Muscle Testing (MMT) scores \cite{cuthbert2007reliability}. Notably, Subject 4 had a low BBS score due to multiple medical conditions but retained moderate lower-limb muscle strength (MMT: 3–4). Despite relying heavily on DRBA’s straps and handles for body-weight support, the reduction in balance demands allowed her lower-limb muscles to function more effectively, contributing to gait improvement. A similar pattern was observed in Subject 3, whose BBS score was close to the threshold for severe balance impairment but who retained MMT scores above 4 in all lower-limb muscle groups. This subject achieved the greatest improvement among all participants. In contrast, Subject 6, whose lower-limb muscle strength ranged from MMT 2- to 4, exhibited limited improvement due to severe ankle dorsiflexion deficits. Even with DRBA’s body-weight support, these functional impairments restricted walking performance.

Qualitative observations indicated that most subjects adapted to DRBA within a few sessions under therapist facilitation. After adaptation, they could walk with DRBA while requiring only occasional therapist intervention. Therapists also reported reduced supervision burden, as the perceived safety and support from DRBA allowed patients to participate in more independent and engaging training.

Although the quantitative analysis focused on walking performance, the sit-to-stand subsystem was also used during patient transfer and preparation for gait training. During the pilot trials, the mechanism assisted seated-to-standing transitions without mechanical failure or safety incidents. Therapists reported that the integrated seat and lifting mechanism reduced physical effort during transfer and supported smoother transition from seated mobility to walking training. A systematic quantitative evaluation of sit-to-stand assistance, including success rate, user comfort, and therapist workload, will be conducted in future work.

Overall, the patient trials show that DRBA can provide immediate improvements in gait performance through adaptive body-weight support and balance assistance. Preliminary observations further indicated that some patients reached 70–80\% of their estimated maximum heart rate during DRBA-assisted walking, suggesting potential for higher-intensity rehabilitation training. However, this study evaluated only immediate assistive effects. Longitudinal clinical studies are required to determine whether these short-term improvements translate into sustained functional recovery.

\section{CONCLUSIONS AND FUTURE WORK}
This paper presented the Dynamic Robotic Balance Assistant (DRBA), a rehabilitation-oriented robotic platform designed to provide assist-as-needed body-weight and balance support for overground gait training. By integrating a compliant pelvic interface, sit-to-stand assistance, and a user-following and fall-intervention architecture, DRBA aims to provide safe, transparent, and adaptable support while preserving natural pelvic motion during rehabilitation.

A preliminary transparency evaluation suggested limited interference with lower-limb kinematics during walking, while pilot studies involving nine elderly patients with balance impairments demonstrated significant improvements in step length and walking speed during DRBA-assisted walking compared with therapist-assisted walking. Furthermore, DRBA enabled users to perform rehabilitation activities beyond their unaided physical capacity and supported higher-intensity training, with some participants reaching 70–80\% of their estimated maximum heart rate during training. These findings demonstrate the immediate assistive benefits of DRBA and highlight its potential to improve rehabilitation outcomes through increased training dosage, engagement, and task-oriented practice.

However, the present study evaluated only the immediate assistive effects of DRBA rather than long-term therapeutic outcomes. Future work will therefore focus on larger transparency evaluations, longitudinal clinical studies to determine whether the increased training opportunities enabled by DRBA translate into sustained functional recovery, and usability studies assessing user comfort, therapist workload, safety, and the effectiveness of the sit-to-stand subsystem.

\addtolength{\textheight}{-1cm}   
\section*{ACKNOWLEDGMENT}
This research is supported by the Singapore National Robotics Programme (NRP) BAU grant - Mobile Robotic Balance Assistant (Award No: M23NBK0045). The authors extend their gratitude to Shiwei Wu, Zhi Wang, Yushu Wang, Chen Tian, and Junhua Liang from the Guangdong Zhongxin Intelligent Rehabilitation Research Institute for their valuable technical assistance and support in experiment setup.


\bibliographystyle{IEEEtran}
\bibliography{Ref}
\end{document}